\documentclass{article}

\usepackage[preprint]{neurips_2024}
\usepackage{dirtytalk}
\usepackage{url}
\usepackage{xcolor}
\usepackage{tikz}
\usepackage{pgfplots}
\usepackage{amsmath}
\usepackage{graphicx}
\usepackage{import}

\usepackage[utf8]{inputenc} 
\usepackage[T1]{fontenc}    
\usepackage{hyperref}       
\usepackage{url}            
\usepackage{booktabs}       
\usepackage{amsfonts}       
\usepackage{nicefrac}       
\usepackage{microtype}      
\usepackage{xcolor}         

\title{Deep Learning-based Prediction of Clinical Trial Enrollment with Uncertainty Estimates}

\author{%
  Tien~Huu~Do\thanks{Equal contribution}\\
  Pfizer \\
  \And
  Antoine~Masquelier\footnotemark[1] \\
  Pfizer \\
  \AND
  Nae Eoun Lee \\
  Pfizer \\
  \And
  Jonathan Crowther\thanks{Work performed while at Pfizer} \\
  Merck \\
}

\begin{document}

\maketitle

\begin{abstract}
            Clinical trials are a systematic endeavor to assess the safety and efficacy of new drugs or treatments.
            Conducting such trials typically demands significant financial investment and meticulous planning, highlighting the need for accurate predictions of trial outcomes. 
            Accurately predicting patient enrollment, a key factor in trial success, is one of the primary challenges during the planning phase. 
            In this work, we propose a novel deep learning-based method to address this critical challenge. 
            Our method, implemented as a neural network model, leverages pre-trained language models (PLMs) to capture the complexities and nuances of clinical documents, transforming them into expressive representations. These representations are then combined with encoded tabular features via an attention mechanism. 
            To account for uncertainties in enrollment prediction, 
            we enhance the model with a probabilistic layer based on the Gamma distribution, 
            which enables range estimation. 
            We apply the proposed model to predict clinical trial duration, assuming site-level enrollment follows a Poisson-Gamma process. 
            We carry out extensive experiments on real-world clinical trial data, and show that the proposed method can effectively predict the number of patients enrolled at a number of sites for a given clinical trial, outperforming established baseline models. 

\end{abstract}

    \section{Introduction}
        \label{sec:intro}
        Drug development is a complex process with multiple stages, including drug discovery, pre-clinical research, clinical research, review, and post-market safety monitoring.
        Each of these stages may include multiple steps. For instance, the clinical research evaluates potential treatments in different phases, which are referred to as~\textit{clinical trials}. 
        Depending on the disease that is being addressed, a clinical trial can span months or even years, involving human participants. Additionally, clinical trials may be conducted in various sites distributed in multiple countries. 
        Given the complexity of clinical trials, substantial investments and efforts are required for their successful execution, rendering the necessity of careful design of the trials. 
        Even with meticulous planning, many clinical trials have failed to produce anticipated clinical outcomes. 
        Recent analyses on different clinical indications have revealed that the success rate of clinical trials is approximately $7.9$\%, which is a high risk for pharmaceutical companies~\cite{kim2023factors}.
        One of the primary reasons for the failure of clinical trials is the insufficient enrollment of patients. Research shows that $19$\% of the trials are terminated due to insufficient enrollment~\cite{carlisle2015unsuccessful} and $80$\% of the trials do not meet initial enrollment goals, resulting in significant financial loss of up to \$$8$ million in revenue per day~\cite{johnson2015evidence}. 
        Therefore, it is of utmost importance to be able to predict the clinical trial success in terms of enrollment before the trials start. This work focuses on enrollment prediction at planning phase.

        Many factors may affect the enrollment outcome of clinical trials such as therapeutic area, patient population, phase, length of treatment, and the geographical distribution of  clinical trial sites. For instance, it is expected that a Vaccine trial has hundreds to thousands of patients while a rare disease trial often has much smaller scale (i.e., tens of patients). Additionally, multi-center trials recruiting patients in numerous countries are likely able to enroll more patients compared to single-center trials. 
        In other words, clinical trial design characteristics provide early indicators of potential enrollment success and can be leveraged for predictive modeling. 

        Many approaches have been proposed for clinical trial enrollment prediction, and they usually fall into two categories: \emph{deterministic} and \emph{stochastic}   approaches.
        The former addresses point estimation while the latter  aims at giving an estimation with some levels of uncertainty. 
        Early deterministic works rely on fixed enrollment rates derived from historical clinical trials~\cite{comfort2013improving}. However, given the non-linear dependencies of the enrollment process on a large number of influential factors, the prediction of enrollment using this approach does not align with reality in most cases. 
        Recent deterministic works leverage advanced machine learning (ML) techniques, producing promising results~\cite{liu2021machine, bieganek2022prediction, amiridi2023enrollment, yue2024trialenroll, yue2024trialdura}. The ML-based methods can model complex relationships and learn from large datasets, enabling more realistic and data-driven prediction thanks to the advance in ML algorithms and innovated hardware capabilities. However, most ML models still rely heavily on structured features and cannot fully capture the nuanced information contained in unstructured clinical text, such as inclusion and exclusion criteria.
        On the other hand, stochastic approach leverages statistical modeling by assuming the observed outcomes follow a single distribution or a mixture of distributions. 
        Although theoretically sound, the stochastic approach often 
        struggles with scalability due to the large volume and heterogeneity of clinical trial data
        ~\cite{carter2004application, carter2005practical, anisimov2008using, anisimov2009predictive, zhong2024enrollment}.

        Our work takes a further step by combining two approaches in a unified model: we propose a novel learning architecture capable of effectively predicting patient enrollment and accounting for prediction uncertainties.
        Our main contributions are as follows:
        \begin{itemize}
            \item We propose a novel model capable of effectively predicting patient enrollment for a given clinical trial design using both structured attributes and unstructured text.  
            The model leverages pre-trained language models (PLMs) to encode textual information and integrates it with structured data using a multi-head attention mechanism, obtaining an expressive representation of the input trial. 
            Subsequently, the representation is used for the downstream task of patient enrollment prediction.
            \item To address uncertainties, we have incorporated a probabilistic component into the proposed architecture through the use of the Gamma distribution. The model learns to predict the parameters of this distribution, and hence the prediction uncertainty can be measured with a confidence interval. By doing so, we demonstrate the flexibility of the proposed method in both settings of point and range estimation.
            \item We apply the proposed stochastic model to predict the duration of clinical trials, leveraging the Poisson-Gamma mixture process. We show that the proposed model can be used in different trial-related contexts.
            \item We empirically show the superior performance of the proposed model compared to strong baselines via extensive experiments and analyses on a large-scale dataset.
        \end{itemize}

        The rest of the paper is organized as follows. Section~\ref{sec:related} positions our work concerning existing works. Section~\ref{sec:method} describes the proposed method in detail. The experiments and discussions are presented in Section~\ref{sec:exp} and we draw our conclusions in Section~\ref{sec:conclusion}.

    \section{Related Work}
        \label{sec:related}
        \subsection{Enrollment Modeling}
            Given potential benefit of enrollment prediction, substantial efforts have been dedicated to the modeling of the patient enrollment. Existing works can be broadly categorized into two approaches: deterministic and stochastic. 
            The deterministic approach focuses on point prediction, relying on fixed enrollment rates or ML algorithms.
            In contrast, the stochastic approach looks at the uncertainty of the enrollment process and assumes that the enrollment follows some prior distributions. The prediction is then produced through simulation, such as the Monte Carlo method~\cite{harrison2010introduction}. 

            Early deterministic methods rely on the assumption of fixed enrollment rates (i.e., patient per site per month) and site initiation rates~\cite{carter2005practical, comfort2013improving} to predict trial duration. More precisely, these methods assume a linear or quadratic relationship between the patient enrollment duration and the total number of patients, enabling the possibility of calculating the enrollment duration based on closed-form solutions. While these method are straightforward to implement, relying on fixed rates seem to be a strong assumption. In addition, estimating the fixed rates from historical data is another challenge given the complexity of clinical trials and the lack of a reliable estimation method. 
            Our work is different from these methods in that we focus on predicting the total number of patients instead of estimating the time needed for the enrollment. In addition, we do not rely on a pre-defined relationship. We instead learn the relationship between number of patients and the characteristics of input trial from a big dataset leveraging the power of advanced deep learning techniques.
            
            The recent advancement in AI and machine learning has given the rise to a multitude of methods for patient enrollment modeling. In~\cite{liu2021machine, bieganek2022prediction}, gradient boosting based models are used to predict number of enrolled patients and enrollment rate at trial-level. Different from these methods, in~\cite{amiridi2023enrollment}, enrollment rate prediction is made via a tensor factorization technique with promising results. 
            However, the tensor factorization method is similar to gradient boosting methods in that they both require extensive feature engineering and hence they struggle to process unstructured clinical text, which plays a critical role in determining enrollment feasibility. Wang \emph{et al.}~\cite{wang2022trial2vec} proposed \textit{trial2vec}, based on BioBERT~\cite{lee2020biobert}, to generate general-purpose embeddings for clinical trials, which can be used for various downstream tasks. Leveraging \textit{trial2vec}, latent topics from complex clinical documents can be found for trial clustering. The found clusters are then used to formulate a sequence feeding a deep recurrent model for predicting trial enrollment outcome~\cite{wang2023spot}. BioBERT is also used in~\cite{yue2024trialdura, yue2024trialenroll} to generate embeddings from eligibility criteria and trial descriptions for downstream predictive tasks. 
            Notably, large language models (LLMs), models with emergent capabilities of generalization to diverse tasks, are leveraged in~\cite{yue2024trialdura} to capture contextual information about trial drug candidates. 
            Similar to these language model based methods, 
            our method relies on deep learning and uses a pre-trained language model, namely Longformer~\cite{li2022clinical}, to generate clinical text embeddings. 
            It is worth noting that we select Longformer owing to its longer context window compared to BERT-based models. 
            In addition, we generate the embeddings on serialized text, i.e., the text created by concatenating multiple attributes of a clinical trial. 
            By doing so, we are able to preserve representativeness and learn contextual embeddings tailored to the prediction task; this differentiates our method from prior approaches that aggregate sentence-level embeddings or simply truncate text. 
            Lastly, we enhance our method by utilizing an extended set of features, and we use the multi-head attention mechanism to effectively combine multi-modal embeddings.

            On the other hand, there exist patient enrollment modeling works that account for the uncertainties of recruitment process. Broadly speaking, methods of this approach assume that the enrollment of patients follows some random distributions, e.g., Poisson distribution. The parameters of these distributions are then found using historical data, and the enrollment scenarios are generated through simulation, e.g., Monte Carlo simulation. In~\cite{carter2004application}, the arrivals of patients within a trial are modeled with a Poisson process. The empirical distribution of trial duration is then found via simulation. While this method is simple and capable of giving a confidence level, it requires a fixed enrollment rate estimated from historical data. However, there are no evidences on how accurate the rate is. 
            Extending this method, a Poisson-Gamma process is used to model the patient enrollment in~\cite{anisimov2009predictive, anisimov2008using, anisimov2007design}, where the site-level enrollment rates are samples from a Gamma distribution. 
            More recently, a Poisson regression model is used to predict the enrollment duration and number of patients at site and trial levels~\cite{zhong2024enrollment}.
            Although results are promising, the method relies on a linear relationship between therapeutic area, trial duration, and some other randomness effects, which might be suboptimal given the complexity of clinical trials.
            Similar to these methods, our approach models the uncertainty of enrollment outcomes using a Gamma distribution. Nevertheless, we learn the parameters of the distribution directly from data leveraging an advanced deep learning algorithm, offering both scalability and generalization across diverse trial types.

        \subsection{Large Language Models}
            Large language models (LLMs) refer to state-of-the-art models with the capabilities of dealing with complex language understanding tasks such as question answering, machine translation, summarization, and text generation. 
            LLMs typically possess an extensive number of parameters and necessitate huge computational resources for training and inference. 
            The popularity of LLMs can be dated back to the introduction of GPT-3~\cite{brown2020language} in $2022$ when end-users were able to interact with LLMs via a chat interface. Since then, many LLMs including LLama~\cite{dubey2024llama}, Gemini~\cite{team2024gemini}, and Claude~\cite{kevian2024capabilities} have been introduced with emergent capabilities of solving diverse reasoning tasks and processing multi-modality data.
            
            In the medical domain, given the complexity of clinical documents, there is an utmost need of understanding  these documents for downstream tasks. There exist approaches that rely on early pre-trained language models (PLMs) such as BioBERT~\cite{lee2020biobert} and ClinicalBERT~\cite{huang2019clinicalbert} for medical text understanding. In these works, BERT~\cite{devlin2019bert} is adapted to specific domains, namely clinical or biomedical domains. 
            Similarly, our work leverages the expressiveness power of PLMs, namely the Longformer model~\cite{li2022clinical} to obtain domain-specific embeddings.  
            However, we do not rely on fine-tuning directly Longformer for downstream tasks. Instead, the embeddings generated by the model are fed into a deep neural network specially designed for patient enrollment prediction. 
            Recent works have leveraged emergent abilities of advanced LLMs for clinical text understanding~\cite{savage2024fine, li2024llamacare}.
            While these works have shown positive results on some tasks such as classification and summarization, little attention has been paid to regression task. 
            In this work, we employ LLM-based fine-tuning as a baseline using Llama2~\cite{touvron2023llama} and show empirically that the proposed method has superior performance compared the LLM fine-tuning based approach on enrollment prediction task.

    \section{Method}
        \label{sec:method}
        \subsection{Problem Statement}
            Denote a clinical trial by $C_i$, our goal is to predict the total number of patients enrolled to the trial $y_i \in \mathbb{N}$. 
            This prediction task can be formalized as a supervised regression problem
            \begin{equation}
                \hat{y}_i = \mathcal{F}(C_i; \Theta),
            \end{equation}
            where $\mathcal{F}$ is the model parameterized by $\Theta$; $\mathcal{F}$ can be thought of as a deterministic function. 
            One might expect the variation of the enrollment given that clinical trials are often complex, involving many factors and random events as discussed in Section~\ref{sec:intro}. 
            Therefore, sometimes it is desirable to account for uncertainties with a certain level of confidence instead of point estimation. Towards this goal, we can modify model $\mathcal{F}$ to yield a distribution of the number of patients: 
            \begin{align}
                \Phi_i &= \mathcal{F}(C_i; \Theta)\\
                Y_i &\sim \mathcal{D}(\Phi_i),
            \end{align}
            where $Y_i$ is a random variable representing number of patients and $\mathcal{D}(\Phi_i)$ is a certain distribution parameterized by $\Phi_i$.

        \subsection{Deterministic Modeling}
        \label{sec:method:deterministic}
            Our first approach is based on a deep neural network model, specially designed for the enrollment regression problem. This approach is referred to as \textit{deterministic modeling} since the proposed model produces a single value given one input trial.

            As discussed in Section~\ref{sec:intro}, a clinical trial is generally complex, and has multiple attributes of miscellaneous types (see Table~\ref{table:data_attribute}). 
            In order to handle these attributes, similar to~\cite{wang2022trial2vec}, we first group the set of trial attributes into two subsets, namely \emph{Key} and \emph{Context}. The~\emph{Key} set contains important attributes that allow the positioning of the trial of interest among reference clinical trials. These attributes are: phase, country, therapeutic area (TA), sponsor, planned number of participants, and planned number of sites. We then apply pre-processing operations to these attributes using MultiLabelBinarizer\footnote{\url{https://scikit-learn.org/stable/modules/generated/sklearn.preprocessing.MultiLabelBinarizer.html}} and standard  transformation (a.k.a. Z-score) to obtain two embeddings $\mathbf{x}^{cat} \in \mathbb{R}^{d_1}$ and $\mathbf{x}^{num} \in \mathbb{R}^{d_2}$. 
            Conversely, the~\emph{Context} set is a mixture of textual and categorical attributes such as title, objective, mechanism of action, indication (disease), inclusion, and exclusion criteria. 
            To effectively process this set, we first 
            apply a simple text serialization technique~\cite{fang2024large} to form a unified text from attributes, that is we concatenate these attributes using a separator. 
            We then generate an embedding $\mathbf{x}^{emb} \in \mathbb{R}^{d_3}$ for the unified text using \textit{Clinical Longformer}~\cite{li2022clinical}. We select Clinical Longformer because it has the context length of $4096$ tokens, which eliminates the issue of context being truncated given that the concatenated text may have thousands of tokens. 
            More importantly, this model has been fine-tuned on clinical corpora, which is capable of capturing nuances in clinical text.

            \begin{figure}[h]
                \centering
                \usetikzlibrary{shapes,arrows}

\definecolor{modelInputColor}{rgb}{.65,0.8,0.99}
\definecolor{NNColor}{rgb}{.99,0.85,0.7}
\definecolor{NNLinkColor}{rgb}{.4,0.3,0.1}
\definecolor{AttentionColor}{rgb}{.8,0.9,0.8}
\definecolor{AddNormColor}{rgb}{.95,0.8,0.95}

\tikzstyle{sum} = [draw, fill=white, circle, minimum width=5mm, inner sep=0pt]
\tikzstyle{neuron} = [draw, fill=white, circle, node distance=1cm, minimum size = 0.2cm, inner sep=0pt]
\tikzstyle{input} = [coordinate, node distance=.5cm]
\tikzstyle{block} = [draw, fill=modelInputColor, rectangle, minimum height=1em, minimum width=1em, text width=1.5cm, align=center, font=\tiny\linespread{0.8}\selectfont, rounded corners=3pt]
\tikzstyle{pinstyle} = [pin edge={to-,thin,black}]

\tikzstyle{NeuralNetBox} = [draw, fill=NNColor, rectangle, minimum height=3.8em, minimum width=2em, text width=1.5cm, align=center, rounded corners=3pt]
\tikzstyle{OutputBox} = [draw, fill=NNColor, rectangle, minimum height=1em, minimum width=1em, text width=0.1cm, align=center, rounded corners=3pt]

\tikzset{
    Output/.pic={
        \node [OutputBox]{};
        \node  [neuron] {};
}}

\tikzset{
Probability/.pic={
    \begin{tikzpicture}[
    declare function={gamma(\z)=
    (2.506628274631*sqrt(1/\z) + 0.20888568*(1/\z)^(1.5) + 0.00870357*(1/\z)^(2.5) - (174.2106599*(1/\z)^(3.5))/25920 - (715.6423511*(1/\z)^(4.5))/1244160)*exp((-ln(1/\z)-1)*\z);},
    declare function={gammapdf(\x,\k,\theta) = \x^(\k-1)*exp(-\x/\theta) / (\theta^\k*gamma(\k));}
]
    \begin{axis}[
      axis x line=none,
      axis y line=none,
      domain=-0:20,
      samples=128,
      xticklabels=\empty,
    ]
    \addplot [red!80!black] {gammapdf(x, 2,3)};
  \end{axis}
\end{tikzpicture}
    
}}

\tikzset{
ProbOutput/.pic={
    \node [OutputBox, minimum width=2em]{};
    \node  [neuron] at (-0.15, 0) {};
    \node  [neuron] at (0.15, 0) {};
}}

\tikzset{
Stacking/.pic={

    \node [OutputBox, , minimum width=5em] at (0.1, 0.1){};
    \foreach \i in {0,1,2,3,4}{
      \node (n2\i) [neuron] at (-\i*0.3+2*0.3+0.1,0+0.1) {};
    }
    
    \node [OutputBox, , minimum width=5em] at (-0.1, -0.1){};
    \foreach \i in {0,1,2,3,4}{
      \node (n2\i) [neuron] at (-\i*0.3+2*0.3-0.1,0-0.1) {};
    }

}}

\tikzset{
    pics/NeuralNet/.style args={#1}{
        code={
        \node [NeuralNetBox]{};
    
        \foreach \i in {0,1,2,3,4}{
            \foreach \j in {0,1,2}{
                \draw [NNLinkColor, -] (-\i*0.3+2*0.3, 0.5) -- (-\j*0.3+1*0.3,0.2);
            }
        }
        \foreach \i in {0,1,2,3,4}{
            \foreach \j in {0,1,2}{
                \draw [NNLinkColor, -] (-\i*0.3+2*0.3,-0.1) -- (-\j*0.3+1*0.3,0.2);
            }
        }
        
        \foreach \i in {0,1,2,3,4}{
          \node (n1\i) [neuron] at (-\i*0.3+2*0.3,-0.1) {};
        }
        \foreach \i in {0,1,2}{
          \node (n2\i) [neuron] at (-\i*0.3+1*0.3,0.2) {};
        }
        \foreach \i in {0,1,2,3,4}{
          \node (n3\i) [neuron] at (-\i*0.3+2*0.3,0.5) {};
        }        
        \node [text height=0.5cm, align=center] at (0, -0.28) {#1};
    }
}}

\tikzstyle{NeuralNet} = [NeuralNetPic, fill=NNColor, rectangle, minimum height=1em, minimum width=8em, text width=1.5cm, align=center]

\def\mynode#{\vtop \bgroup \hsize 0pt \parindent 0pt  \rightskip = 0pt minus \maxdimen \let\next=}

\begin{tikzpicture}[auto, node distance=2cm,>=latex']

    \node [block, name=txt_emb] {Text Embeddings};

    \node [input, name=blank1, right of=txt_emb] {};

    \node [block, name=phase, right of=blank1] {Phase};
    \node [block, name=countries, right of=phase] {Countries};
    \node [block, name=ta, right of=countries] {TA};
    \node [block, name=sponsor, right of=ta] {Sponsor};

    \node [input, name=blank2, right of=sponsor] {};

    \node [block, name=PlannedNbSubjects, right of=blank2] {Planned \#Subjects};
    \node [block, name=PlannedNbSites, right of=PlannedNbSubjects] {Planned \#Sites};

    \node [coordinate, name=cat_in, above of=countries, right of=countries, node distance=1cm] {};
    \node [coordinate, name=cat_in2, above of=cat_in, node distance=1cm] {};
    \node [coordinate, name=phase_in, above of=phase, node distance=1cm] {};
    \draw [-] (phase) -- node{} (phase_in);
    \draw [-] (phase_in) -- node{} (cat_in);
    \node [coordinate, name=countries_in, above of=countries, node distance=1cm] {};
    \draw [-] (countries) -- node{} (countries_in);
    \node [coordinate, name=ta_in, above of=ta, node distance=1cm] {};
    \draw [-] (ta) -- node{} (ta_in);
    \node [coordinate, name=sponsor_in, above of=sponsor, node distance=1cm] {};
    \draw [-] (sponsor) -- node{} (sponsor_in);
    \draw [-] (sponsor_in) -- node{} (cat_in);

    \node [coordinate, name=num_in, above of=PlannedNbSubjects, right of=PlannedNbSubjects, node distance=1cm] {};
    \node [coordinate, name=num_in2, above of=num_in, node distance=1cm] {};
    \node [coordinate, name=PlannedNbSubjects_in, above of=PlannedNbSubjects, node distance=1cm] {};
    \draw [-] (PlannedNbSubjects) -- node{} (PlannedNbSubjects_in);
    \draw [-] (PlannedNbSubjects_in) -- node{} (num_in);
    \node [coordinate, name=PlannedNbSites_in, above of=PlannedNbSites, node distance=1cm] {};
    \draw [-] (PlannedNbSites) -- node{} (PlannedNbSites_in);
    \draw [-] (PlannedNbSites_in) -- node{} (num_in);

    \pic [local bounding box=EmbNN, above of=txt_emb, node distance=2.2cm]{NeuralNet={\tiny{$\mathcal{F}_{\textit{emb}}$}}};
    \draw [->] (txt_emb) -- node{\tiny{$\mathbf{x}^{\textit{emb}}$}} (EmbNN);
    \pic [local bounding box=CatNN, above of=cat_in, node distance=1.2cm]{NeuralNet={\tiny{$\mathcal{F}_{\textit{cat}}$}}};
    \draw [->] (cat_in) -- node{\tiny{$\mathbf{x}^{\textit{cat}}$}} (CatNN);
    \pic [local bounding box=NumNN, above of=num_in, node distance=1.2cm]{NeuralNet={\tiny{$\mathcal{F}_{\textit{num}}$}}};
    \draw [->] (num_in) -- node{\tiny{$\mathbf{x}^{\textit{num}}$}} (NumNN);

    \node [coordinate, name=concat_num, above of=NumNN, node distance=1.5cm] {};
    \pic [local bounding box=stacking, above of=CatNN, node distance=1.5cm]{Stacking};
    \draw [->] (CatNN) -- node{\tiny{$\mathbf{z}^{\textit{cat}}$}} (stacking);
    \draw [-] (NumNN) -- node{} (concat_num);
    \draw [->, align=left] (concat_num) -- node{\tiny{$\mathbf{z}^{\textit{num}}$}} (stacking);

    \node [coordinate, name=attention_numcat, above of=stacking, node distance=0.8cm] {};
    \node [coordinate, name=attention_numcat2, right of=attention_numcat, node distance=1.8cm] {};
    \node [coordinate, name=attention_numcat3, above of=attention_numcat2, node distance=1.2cm] {};
    \node [block, name=attention, above of=attention_numcat, node distance=1.2cm, fill=AttentionColor, text width=2.5cm] {\footnotesize Multi-headed Attention};
    \draw [-] (stacking) -- node{} (attention_numcat);
    \draw [->] (attention_numcat) -- node{\tiny{K}} (attention);
    \draw [-] (attention_numcat) -- node{} (attention_numcat2);
    \draw [-] (attention_numcat2) -- node{} (attention_numcat3);
    \draw [->] (attention_numcat3) -- node{\tiny{V}} (attention);

    \node [coordinate, name=attention_emb, above of=txt_emb, node distance=5.7cm] {};
    \node [coordinate, name=attention_emb2, right of=attention_emb, node distance=3.2cm] {};
    \node [coordinate, name=attention_emb3, above of=attention_emb2, node distance=1cm] {};
    \draw [-] (EmbNN) -- node{} (attention_emb);
    \draw [-] (attention_emb) -- node{\tiny{$\mathbf{z}^{\textit{emb}}$}} (attention_emb2);
    \draw [->] (attention_emb2) -- node{\tiny{Q}} (attention);

    \node [block, name=add_norm, above of=attention, node distance=1cm, fill=AddNormColor, text width=2.5cm] {\footnotesize Add \& Norm};
    \draw [-] (attention_emb2) -- node{} (attention_emb3);
    \draw [->] (attention_emb3) -- node{} (add_norm);
    \draw [->] (attention) -- node{} (add_norm);

    \pic [local bounding box=output, above of=add_norm, node distance=1.2cm]{Output};
    \draw [->] (add_norm) -- node{} (output);
    
\label{fig:1}
\end{tikzpicture}
                \caption{Deterministic model architecture for enrollment prediction. Two sets of attributes are \textit{Key} and \textit{Context}, where the latter  is encoded into text \textit{embeddings} using Clinical Longformer. $\mathcal{F}_{emb}, \mathcal{F}_{cat}$, and  $\mathcal{F}_{num}$ are fully connected layers.
                }
                \label{fig:deterministic_model}
            \end{figure}
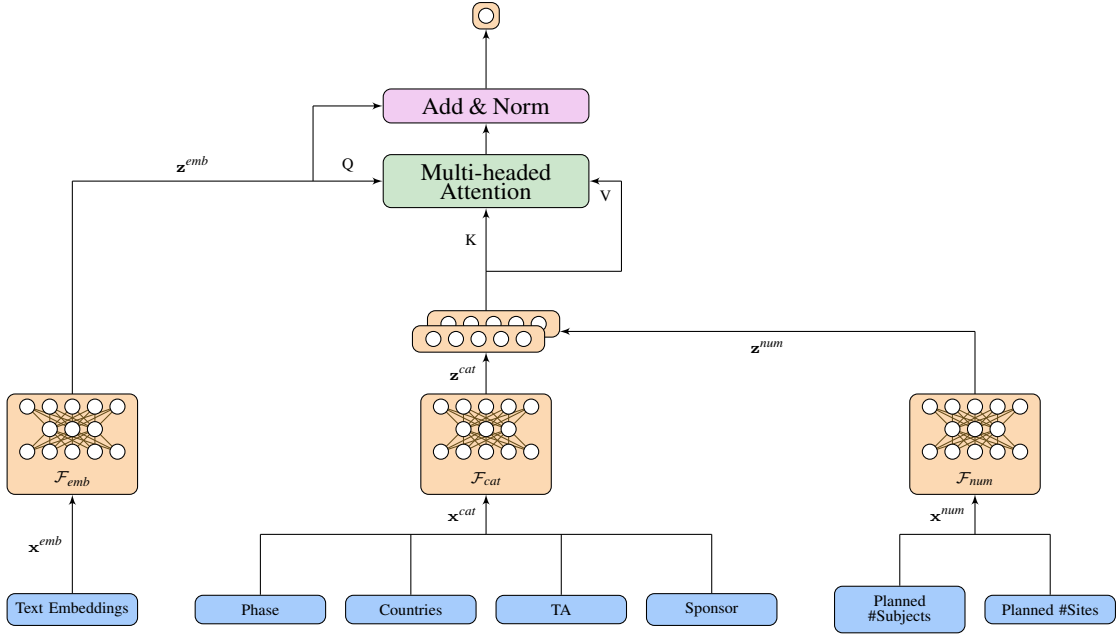


            Our model is presented in~Figure~\ref{fig:deterministic_model}, where it takes as inputs three embeddings $\mathbf{x}^{emb}$, $\mathbf{x}^{cat}$, and $\mathbf{x}^{num}$. Using several fully connected (FC) layers with activations, we obtain intermediate representations:
            \begin{subequations}
            \label{eq:embeddings}
            \begin{align}
                \mathbf{z}^{emb} &= \mathcal{F}_{emb}(\mathbf{x}^{emb}; \Theta_1), \label{eq:txt_embeddings}\\
                \mathbf{z}^{cat} &= \mathcal{F}_{cat}(\mathbf{x}^{cat}; \Theta_2), \label{eq:cat_embeddings}\\
                \mathbf{z}^{num} &= \mathcal{F}_{num}(\mathbf{x}^{num}; \Theta_3), \label{eq:num_embeddings}
            \end{align}
            \end{subequations}
            where $\mathbf{z}^{emb}, \mathbf{z}^{cat}, \mathbf{z}^{num} \in \mathbb{R}^D$. 
            These vectors are then combined using the multi-head attention mechanism~\cite{waswani2017attention} where
            \begin{align}
                Q &= \text{unsqueeze}_1(\mathbf{z}^{emb})\\ 
                K &= V = (\mathbf{z}^{cat}, \mathbf{z}^{num}) 
            \end{align}

            with $\text{unsqueeze}_1 : \mathbb{R}^{B \times 32} \rightarrow \mathbb{R}^{B \times 1 \times d_3}$ and $B$ being the batch size.\\
            
             This operation yields  $\mathbf{z}^{att} = \text{MultiHead}(Q, K, V) \in \mathbb{R}^D$. In the end, we add a skip connection~\cite{he2016deep} and a layer-norm operation~\cite{ba2016layer} to produce the final representation:
            \begin{equation}
                \mathbf{h} = \text{LayerNorm}( \mathbf{z}^{att} + \mathbf{z}^{emb})
            \end{equation}
            This final embedding $\mathbf{h}$ is passed to a regression head to produce the predicted number of patients.

             We train the proposed model by minimizing a $L_1$ loss, defined by:
             \begin{equation}
                 \mathcal{L}(\Theta) =  \frac{1}{n}\sum _{i=1}^{n} \left| \text{ln}\ y_{i} - \text{ln}\ \hat{y_{i}} \right|,
             \end{equation}
             where $n$ is the number of trials in our dataset. 
             It is worth noting that $\mathcal{L}(\Theta)$ is calculated on logarithmic scale since we found empirically that utilizing the logarithmic scale loss function brings better prediction performance.
    
        \subsection{Stochastic Modeling}
        \label{sec:method:stochastic}
\begin{figure}
                    \centering
                    \includegraphics[scale=0.41]{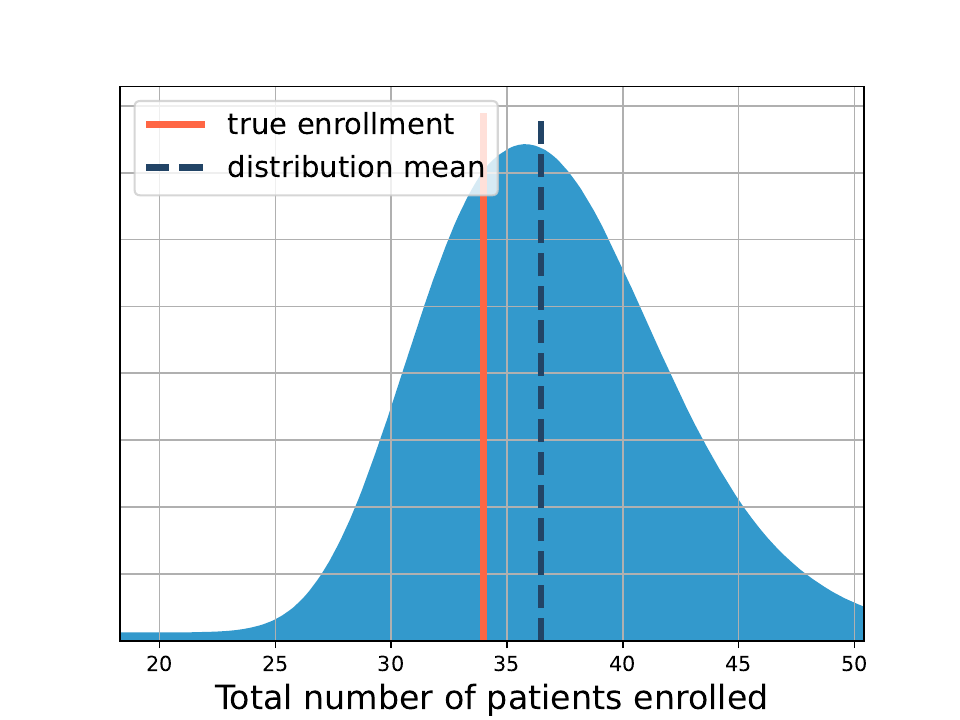}
                    \includegraphics[scale=0.41]{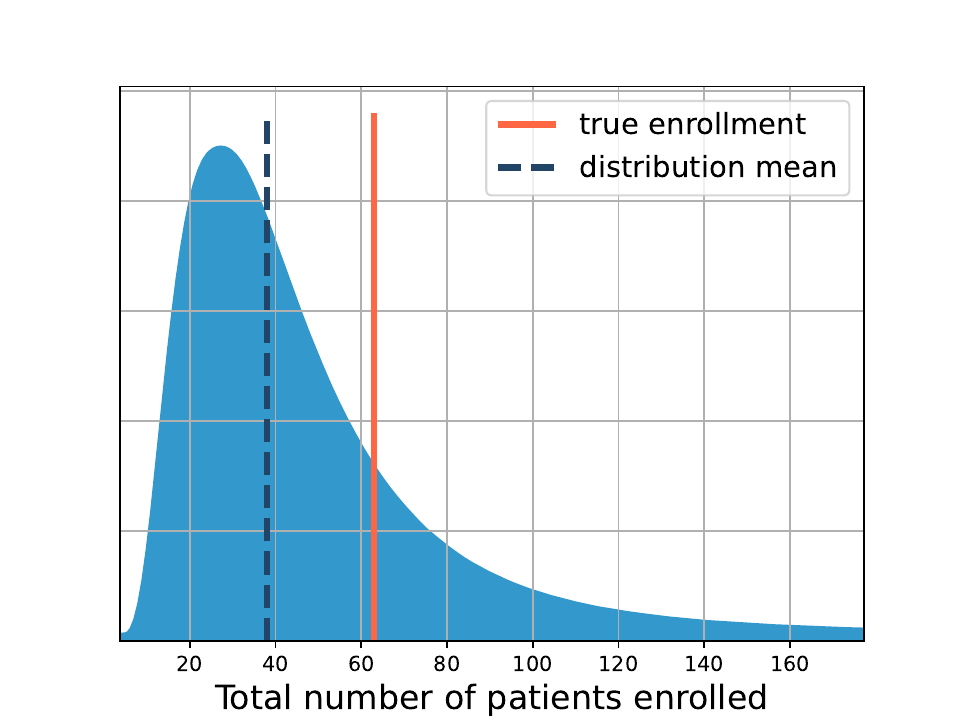}
                    \caption{Examples of output of the stochastic model for two different studies. The dashed line represents the mean of the output distribution and the orange line represents the true total number of patients enrolled. The horizontal axis represents the number of patients enrolled and the vertical axis represents its probability density.}
                    \label{fig:stochastic_output}
            \end{figure}

            It is desirable to design clinical trials with predictable outcomes, e.g., in terms of patient enrollment. However, clinical trials in reality often include some sorts of uncertainty due to factors like patient availability, site activation delays, and external disruptions. 
            In order to account for uncertainties, we extend the proposed deterministic model in Section~\ref{sec:method:deterministic} by predicting the distribution of the number of enrolled patients instead of a single number. More precisely, the model predicts the parameters of the distribution, which in our case is chosen to be a Gamma distribution due to its flexibility to fit a wide variety of shapes for right-skewed data. This way, the model can not only predict the number of participants via sampling but it can also give a range estimation with a confidence level. Examples of such probabilistic outputs are shown in Figure~\ref{fig:stochastic_output}.

\begin{figure}[h]
                \centering
                \usetikzlibrary{shapes,arrows}

\definecolor{modelInputColor}{rgb}{.65,0.8,0.99}
\definecolor{NNColor}{rgb}{.99,0.85,0.7}
\definecolor{NNLinkColor}{rgb}{.4,0.3,0.1}
\definecolor{AttentionColor}{rgb}{.8,0.9,0.8}
\definecolor{AddNormColor}{rgb}{.95,0.8,0.95}

\tikzstyle{sum} = [draw, fill=white, circle, minimum width=5mm, inner sep=0pt]
\tikzstyle{neuron} = [draw, fill=white, circle, node distance=1cm, minimum size = 0.2cm, inner sep=0pt]
\tikzstyle{input} = [coordinate, node distance=.5cm]
\tikzstyle{block} = [draw, fill=modelInputColor, rectangle, minimum height=1em, minimum width=1em, text width=1.5cm, align=center, font=\tiny\linespread{0.8}\selectfont, rounded corners=3pt]
\tikzstyle{pinstyle} = [pin edge={to-,thin,black}]

\tikzstyle{NeuralNetBox} = [draw, fill=NNColor, rectangle, minimum height=3.8em, minimum width=2em, text width=1.5cm, align=center, rounded corners=3pt]
\tikzstyle{OutputBox} = [draw, fill=NNColor, rectangle, minimum height=1em, minimum width=1em, text width=0.1cm, align=center, rounded corners=3pt]

\tikzset{
    Output/.pic={
        \node [OutputBox]{};
        \node  [neuron] {};
}}

\tikzset{
Probability/.pic={
    \begin{tikzpicture}[
    declare function={gamma(\z)=
    (2.506628274631*sqrt(1/\z) + 0.20888568*(1/\z)^(1.5) + 0.00870357*(1/\z)^(2.5) - (174.2106599*(1/\z)^(3.5))/25920 - (715.6423511*(1/\z)^(4.5))/1244160)*exp((-ln(1/\z)-1)*\z);},
    declare function={gammapdf(\x,\k,\theta) = \x^(\k-1)*exp(-\x/\theta) / (\theta^\k*gamma(\k));}
]
    \begin{axis}[
      axis x line=none,
      axis y line=none,
      domain=-0:20,
      samples=128,
      xticklabels=\empty,
    ]
    \addplot [red!80!black] {gammapdf(x, 2,3)};
  \end{axis}
\end{tikzpicture}
    
}}

\tikzset{
ProbOutput/.pic={
    
    \node [OutputBox, minimum width=3.5em, minimum height=3em]{};
    \node [input] at (0,0) {\tiny$\Gamma(\alpha, \lambda)$};
    \node  [neuron, minimum width=1em] at (-0.3,-0.25) {\small $\alpha$};
    \node  [neuron, minimum width=1em] at (0.3,-0.25) {\small $\lambda$};
    \node [text width=1.5cm, align=center] at (0,0.25) {\small$\Gamma(\alpha, \lambda)$};
}}

\tikzset{
Stacking/.pic={

    \node [OutputBox, , minimum width=5em] at (0.1, 0.1){};
    \foreach \i in {0,1,2,3,4}{
      \node (n2\i) [neuron] at (-\i*0.3+2*0.3+0.1,0+0.1) {};
    }
    
    \node [OutputBox, , minimum width=5em] at (-0.1, -0.1){};
    \foreach \i in {0,1,2,3,4}{
      \node (n2\i) [neuron] at (-\i*0.3+2*0.3-0.1,0-0.1) {};
    }

}}

\tikzset{
    pics/NeuralNet/.style args={#1}{
        code={
        \node [NeuralNetBox]{};
    
        \foreach \i in {0,1,2,3,4}{
            \foreach \j in {0,1,2}{
                \draw [NNLinkColor, -] (-\i*0.3+2*0.3, 0.5) -- (-\j*0.3+1*0.3,0.2);
            }
        }
        \foreach \i in {0,1,2,3,4}{
            \foreach \j in {0,1,2}{
                \draw [NNLinkColor, -] (-\i*0.3+2*0.3,-0.1) -- (-\j*0.3+1*0.3,0.2);
            }
        }
        
        \foreach \i in {0,1,2,3,4}{
          \node (n1\i) [neuron] at (-\i*0.3+2*0.3,-0.1) {};
        }
        \foreach \i in {0,1,2}{
          \node (n2\i) [neuron] at (-\i*0.3+1*0.3,0.2) {};
        }
        \foreach \i in {0,1,2,3,4}{
          \node (n3\i) [neuron] at (-\i*0.3+2*0.3,0.5) {};
        }        
        \node [text height=0.5cm, align=center] at (0, -0.28) {#1};
    }
}}

\tikzstyle{NeuralNet} = [NeuralNetPic, fill=NNColor, rectangle, minimum height=1em, minimum width=8em, text width=1.5cm, align=center]

\def\mynode#{\vtop \bgroup \hsize 0pt \parindent 0pt  \rightskip = 0pt minus \maxdimen \let\next=}

\begin{tikzpicture}[auto, node distance=2cm,>=latex]

    \node [block, name=txt_emb] {Text Embeddings};

    \node [input, name=blank1, right of=txt_emb] {};

    \node [block, name=phase, right of=blank1] {Phase};
    \node [block, name=countries, right of=phase] {Countries};
    \node [block, name=ta, right of=countries] {TA};
    \node [block, name=sponsor, right of=ta] {Sponsor};

    \node [input, name=blank2, right of=sponsor] {};

    \node [block, name=PlannedNbSubjects, right of=blank2] {Planned \#Subjects};
    \node [block, name=PlannedNbSites, right of=PlannedNbSubjects] {Planned \#Sites};

    \node [coordinate, name=cat_in, above of=countries, right of=countries, node distance=1cm] {};
    \node [coordinate, name=cat_in2, above of=cat_in, node distance=1cm] {};
    \node [coordinate, name=phase_in, above of=phase, node distance=1cm] {};
    \draw [-] (phase) -- node{} (phase_in);
    \draw [-] (phase_in) -- node{} (cat_in);
    \node [coordinate, name=countries_in, above of=countries, node distance=1cm] {};
    \draw [-] (countries) -- node{} (countries_in);
    \node [coordinate, name=ta_in, above of=ta, node distance=1cm] {};
    \draw [-] (ta) -- node{} (ta_in);
    \node [coordinate, name=sponsor_in, above of=sponsor, node distance=1cm] {};
    \draw [-] (sponsor) -- node{} (sponsor_in);
    \draw [-] (sponsor_in) -- node{} (cat_in);

    \node [coordinate, name=num_in, above of=PlannedNbSubjects, right of=PlannedNbSubjects, node distance=1cm] {};
    \node [coordinate, name=num_in2, above of=num_in, node distance=1cm] {};
    \node [coordinate, name=PlannedNbSubjects_in, above of=PlannedNbSubjects, node distance=1cm] {};
    \draw [-] (PlannedNbSubjects) -- node{} (PlannedNbSubjects_in);
    \draw [-] (PlannedNbSubjects_in) -- node{} (num_in);
    \node [coordinate, name=PlannedNbSites_in, above of=PlannedNbSites, node distance=1cm] {};
    \draw [-] (PlannedNbSites) -- node{} (PlannedNbSites_in);
    \draw [-] (PlannedNbSites_in) -- node{} (num_in);

    \pic [local bounding box=EmbNN, above of=txt_emb, node distance=2.2cm]{NeuralNet={\tiny{$\mathcal{F}_{\textit{emb}}$}}};
    \draw [->] (txt_emb) -- node{\tiny{$\mathbf{x}^{\textit{emb}}$}} (EmbNN);
    \pic [local bounding box=CatNN, above of=cat_in, node distance=1.2cm]{NeuralNet={\tiny{$\mathcal{F}_{\textit{cat}}$}}};
    \draw [->] (cat_in) -- node{\tiny{$\mathbf{x}^{\textit{cat}}$}} (CatNN);
    \pic [local bounding box=NumNN, above of=num_in, node distance=1.2cm]{NeuralNet={\tiny{$\mathcal{F}_{\textit{num}}$}}};
    \draw [->] (num_in) -- node{\tiny{$\mathbf{x}^{\textit{num}}$}} (NumNN);

    \node [coordinate, name=concat_num, above of=NumNN, node distance=1.5cm] {};
    \pic [local bounding box=stacking, above of=CatNN, node distance=1.5cm]{Stacking};
    \draw [->] (CatNN) -- node{\tiny{$\mathbf{z}^{\textit{cat}}$}} (stacking);
    \draw [-] (NumNN) -- node{} (concat_num);
    \draw [->, align=left] (concat_num) -- node{\tiny{$\mathbf{z}^{\textit{num}}$}} (stacking);

    \node [coordinate, name=attention_numcat, above of=stacking, node distance=0.8cm] {};
    \node [coordinate, name=attention_numcat2, right of=attention_numcat, node distance=1.8cm] {};
    \node [coordinate, name=attention_numcat3, above of=attention_numcat2, node distance=1.2cm] {};
    \node [block, name=attention, above of=attention_numcat, node distance=1.2cm, fill=AttentionColor, text width=2.5cm] {\footnotesize Multi-headed Attention};
    \draw [-] (stacking) -- node{} (attention_numcat);
    \draw [->] (attention_numcat) -- node{\tiny{K}} (attention);
    \draw [-] (attention_numcat) -- node{} (attention_numcat2);
    \draw [-] (attention_numcat2) -- node{} (attention_numcat3);
    \draw [->] (attention_numcat3) -- node{\tiny{V}} (attention);

    \node [coordinate, name=attention_emb, above of=txt_emb, node distance=5.7cm] {};
    \node [coordinate, name=attention_emb2, right of=attention_emb, node distance=3.2cm] {};
    \node [coordinate, name=attention_emb3, above of=attention_emb2, node distance=1cm] {};
    \draw [-] (EmbNN) -- node{} (attention_emb);
    \draw [-] (attention_emb) -- node{\tiny{$\mathbf{z}^{\textit{emb}}$}} (attention_emb2);
    \draw [->] (attention_emb2) -- node{\tiny{Q}} (attention);

    \node [block, name=add_norm, above of=attention, node distance=1cm, fill=AddNormColor, text width=2.5cm] {\footnotesize Add \& Norm};
    \draw [-] (attention_emb2) -- node{} (attention_emb3);
    \draw [->] (attention_emb3) -- node{} (add_norm);
    \draw [->] (attention) -- node{} (add_norm);

    \pic [local bounding box=output, above of=add_norm, node distance=1.2cm]{ProbOutput};
    \draw [->] (add_norm) -- node{} (output);
    
\label{fig:2}
\end{tikzpicture}
                \caption{Stochastic model architecture for patient enrollment prediction. This model shares the architecture with the deterministic model except the last layer where two parameters of the Gamma distribution are predicted.}
                \label{fig:probabilistic_model}
            \end{figure}
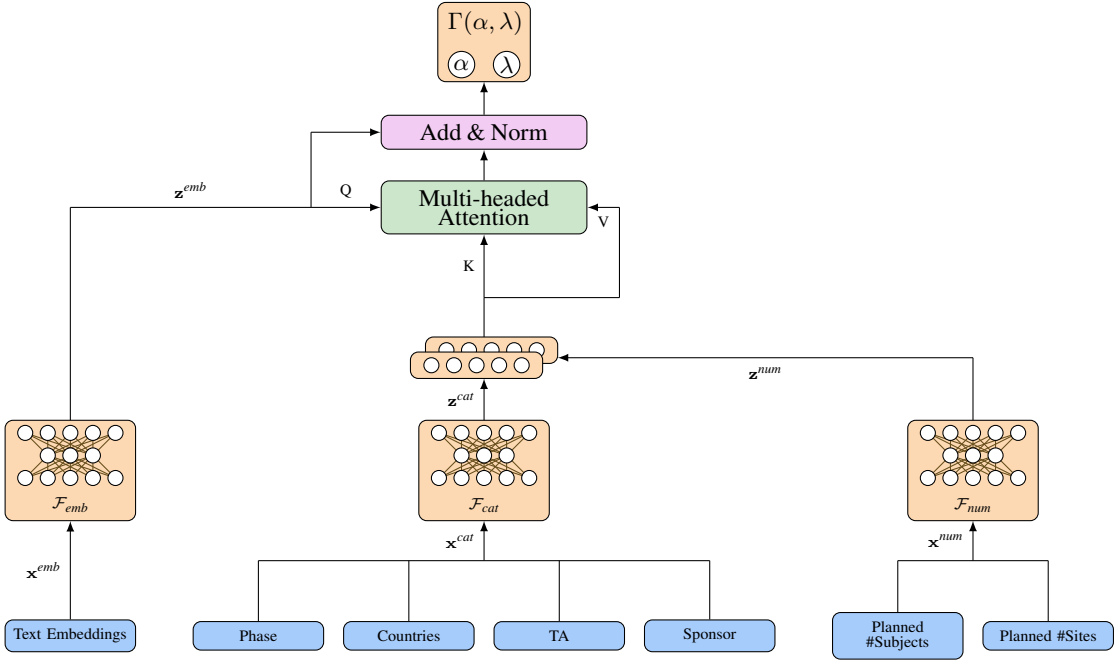

            The stochastic model, depicted in Figure~\ref{fig:probabilistic_model}, employs the same backbone as the architecture outlined in Section~\ref{sec:method:deterministic}, but its final layer branches into two outputs that parametrize the Gamma distribution. The logits are passed through an exponential function to ensure that the shape $\alpha$ and rate $\lambda$ are strictly positive
            \begin{align}
                \alpha &= \exp(\text{FC}_{\alpha}(\mathbf{h})) \\
                \lambda &= \exp(\text{FC}_{\lambda}(\mathbf{h})) ,
            \end{align}
            where FC refers to a fully connected layer with a Leaky ReLU function applied at the end of each layer.\\

            Since the PDF of the distribution is known, the loss can be defined by:
            \begin{equation}
                 \mathcal{L}(\Theta) =  -\frac{1}{n}\sum _{i=1}^{n} \text{ln}\ p(\text{ln}\ \hat{y_{i}}\ |\ C_i)
             \end{equation}
        
        \subsection{Deep Poisson-Gamma}
        \label{sec:method:poissongamma}
            The stochastic approach described in Section~\ref{sec:method:stochastic} is more challenging to train due to its variable nature. 
            However, its ability to capture variability can be beneficial in cases where a same study design leads to a wide variety of outcomes. 
            Let us consider the Poisson-Gamma enrollment forecasting framework proposed by Anisimov~\cite{anisimov2008using}, which models the enrollment of a trial as a mixture of Poisson processes, where each Poisson process represents the enrollment of a site. The rates of these Poisson processes follow a Gamma distribution $\Gamma_{\mu}(\alpha, \lambda)$. Additionally, for each site, the Poisson process starts at a random startup time following a distribution $\Gamma_{\theta}(\alpha, \lambda)$. 
            In this particular use case, a same input leads to a very wide range of rates and startup times combinations as we are predicting site-level output based on trial-level information. The work suggests that, at planning stage, the parameters $\alpha$ and $\lambda$ of the enrollment rates can be evaluated by using historical data from similar trials and that the startup times could be modeled as a random variable. Inspired by Anisimov's work, we explore two different approaches to evaluate the parameters of the Poisson-Gamma model that are used to predict the enrollment \emph{duration} of a trial at planning phase based on its attributes. 
            
            We model the enrollment process of a trial $C_i$ with a set of sites $S_i$ as a mixture of Poisson processes as described in Anisimov~\cite{anisimov2008using}. The recruitment $\rho_{i,s,t}$ of each site $s \in S_i$ at time $t$ is modeled as a simple Poisson process with a rate $\mu_{i,s}$ and a startup time $\theta_{i,s}$. The \textit{Target enrollment} from $C_i$ is also used in the modeling and will be denoted as $\pi_i$.  \\

            The model is defined by
            \begin{equation} \label{eq:poisson_gamma:rho}
                \rho_{i,s,t}= 
                   \begin{cases}
                       \sim Poisson(\mu_{i,s})    & \text{if } t \geq \theta_{i,s}\\
                       0                           & \text{otherwise}
                   \end{cases}
                \\
            \end{equation}
            \begin{equation} \label{eq:poisson_gamma:R}
                R_{i,T} = \sum\limits_{t=0}^{T} \sum\limits_{s \in S_i}^{}  \rho_{i,s,t}
            \end{equation}
            \begin{equation} \label{eq:poisson_gamma:tau}
                \hat{\tau}_i = min\{ t \in \mathbb{Z}\ |\ R_{i,t} \geq \pi_i\},
            \end{equation}\\
            where $\rho_{i,s,t}$ is the number of enrolled patients of a site $s$ at time $t$, $R_{i,T}$ the total number of patients for trial $C_i$ at time $T$. The predicted enrollment duration of a trial $\hat{\tau}_i$ is defined as the first time step at which the total enrollment $R_{i,t}$ reaches the target enrollment $\pi_i$.\\

            
            For the first approach, we train a slightly different version of our stochastic model from Section \ref{sec:method:stochastic}. The outputs are now the parameters of the distributions of the enrollment rate and the startup time of a site $\Gamma_{\mu, i}$ and $\Gamma_{\theta, i}$. These distributions are then described by the following equations:
            \begin{equation} \label{eq:poisson_gamma:Gamma_stocha}
                ((\alpha_{\mu, i}, \lambda_{\mu, i}),(\alpha_{\theta, i}, \lambda_{\theta, i}))  = \mathcal{F}_{(\mu, \theta)}(C_i; \Theta)
            \end{equation}
            \begin{equation} \label{eq:poisson_gamma:Gamma_stocha_mu}
                \Gamma_{\mu, i}  \equiv \Gamma(\alpha_{\mu, i}, \lambda_{\mu, i})\\
            \end{equation}
            \begin{equation} \label{eq:poisson_gamma:Gamma_stocha_theta}
                \Gamma_{\theta, i}  \equiv \Gamma(\alpha_{\theta, i}, \lambda_{\theta, i})\\
            \end{equation}

            The target outputs in our model are vectors of enrollment rates and startup times from sites $S_i$ in their original scale. 
            The architecture used is the same as the one described in Section~\ref{sec:method:stochastic}. However, due to the different nature and scale of the target output, the loss function as described in equations~\eqref{eq:poisson_gamma:loss} have been modified accordingly.

            \begin{equation}\label{eq:poisson_gamma:loss}
                 \mathcal{L_{\mu}}(\Theta) =  -\frac{1}{n}\sum _{i=1}^{n} \frac{1}{|S_i|}\sum _{_{i,s}}^{S_i}   \text{ln}\ p( \hat{\mu}_{i,s}\ |\  C_i)
            \end{equation}\\
            \begin{equation}\label{eq:poisson_gamma:loss}
                 \mathcal{L_{\theta}}(\Theta) =  -\frac{1}{n}\sum _{i=1}^{n} \frac{1}{|S_i|}\sum _{_{i,s}}^{S_i} \text{ln}\ p( \hat{\theta}_{i,s}\ |\  C_i)
            \end{equation}\\
            \begin{equation}\label{eq:poisson_gamma:loss_total}
                 \mathcal{L}(\Theta) =  \mathcal{L_{\mu}}(\Theta) + \mathcal{L_{\theta}}(\Theta)
            \end{equation}\\

            
            

            For the second approach, we implement a filtering-and-fitting baseline inspired by Anisimov~\cite{anisimov2008using} and Zhong \emph{et al.} \cite{zhong2024enrollment}. The method identifies, for each trial $C_i$, historical trials with similar characteristics (based on the categorical features listed in Table~\ref{table:data_attribute}) and fits Gamma distributions $\Gamma_{\mu, i}$ and $\Gamma_{\theta, i}$ to, respectively, their historical site-level enrollment rates and startup times using maximum likelihood estimation (MLE).

            Let the feature set $\mathcal{F}$ include attributes such as phase, country, and therapeutic area. Each feature $f \in \mathcal{F}$ can have multiple values for a given trial. Let $f_i$ be the set of values for feature $f$ in trial $C_i$. We consider trial $C_d$ to be part of the set of similar trials if:\\ 
            \begin{equation}
                \forall f : f_d \cap f_i \neq \emptyset
            \end{equation}\\
            This ensures that the historical trials used to fit the distributions $\Gamma_{\mu,i}$ and $\Gamma_{\theta,i}$ are sufficiently similar in design. 
            


    \section{Experiments}
    \label{sec:exp}
        \subsection{Dataset}
        \label{sec:dataset}
            We collect trial-level data from two primary sources, namely IQVIA Data Query System (DQS\footnote{\url{https://www.iqvia.com/solutions/technologies/orchestrated-clinical-trials/planning-suite/data-query-system}}) and Citeline\footnote{\url{https://www.citeline.com/en}}. 
            DQS contains proprietary operational data about clinical trials, principal investigators, and facilities. 
            The data in DQS are provided by pharmaceutical companies or collected from shared data collaborations such as Investigator Registry or the Investigator Databank. 
            Thanks to its volume and high quality, DQS enables sponsors and contract research organizations (CROs) to find the best matched investigators and facilities, and hence effectively support clinical trial planning.
            Data from Citeline come from more than $40$k sources\footnote{\url{https://www.informa.com/globalassets/documents/investor-relations/2019/investorday/citeline.pdf}} of monitored information, and are curated by their scientists. Similar to DQS, Citeline data contain information about trials, investigators, and facilities. 
            Citeline data lack the detailed site-level performance metrics that DQS offers, however, Citeline provides extra features and metrics at trial and drug levels. 
            Hence, we collect the information about number of sites, trial duration, and number of enrolled patients from DQS while other structured and textual attributes are taken from Citeline. 
            Another data source is \textit{ClinicalTrials.gov}, which is a publicly available database established by National Institute of Health (NIH). 
            It should be noted that a large portion of Citeline trial data can be found in this database. 
            
            Our dataset contains more than $11.4$k clinical trials of different therapeutic areas such as Oncology, Inflammation, and Cardiovascular. 
            Since our method is based on supervised learning, 
            we only consider trials with known outcome. For this reason,  the status of a trial in our dataset must be either \textit{Completed} or \textit{Closed}. 
            The number of enrolled patients may vary greatly across trials depending on the nature of therapeutic areas, ranging between $1$ and $11.7$k. 
            A comprehensive list of attributes along with their corresponding summary descriptions is presented in Table~\ref{table:data_attribute}.

\begin{table*}[t]
    \centering
    \caption{The list of attributes and label in our dataset collected from Citeline and DQS. 
    The target enrollment is the planned number of patients set by clinical trial managers. 
    The number of enrolled patients (No. enrolled) is the actual number of patients when the corresponding trial completes, which is our prediction target. 
    It should be noted that the target enrollment number in some cases is close to the actual enrolled number but in many cases there are wide discrepancies between those numbers. `` - '' stands for attributes that are not applicable to the corresponding columns and TA stands for Therapeutic Area. 
    Categorical attributes may be single-valued or multi-valued; those marked with a dagger are multi-valued. 
    Attribute \emph{Phase} has eight different values because some trials are multi-phase (e.g., II/III).\\
    }
    \label{table:data_attribute}
    \resizebox{\textwidth}{!}{
     \begin{tabular}{c | c | c | c | c | c | c | c} 
     \hline \hline
     \textbf {Attribute} & \textbf{Source} & \textbf {Type} & \textbf{Min} & \textbf{Median} & \textbf{Mean} & \textbf{Max} & \textbf{Cardinality}\\
            \hline
            \hline
            Title & Citeline & Text & - & - & - & - & - \\
            \hline
            Objective & Citeline & Text & - & - & - & - & - \\
            \hline
            TA & Citeline & Categorical & - & - & - & - & $9$\\
            \hline
            Indication\textsuperscript{\textdagger} & Citeline & Categorical & - & - & - & - & $970$\\
            \hline
            Mechanism\textsuperscript{\textdagger} & Citeline & Categorical & - & - & - & - & $995$\\
            \hline
            Sponsor\textsuperscript{\textdagger} & Citeline & Categorical &  - & - & - & - & $962$\\
            \hline
            Sponsor Type\textsuperscript{\textdagger} & Citeline & Categorical &  - & - & - & - & $10$\\
            \hline
            Country\textsuperscript{\textdagger} & Citeline & Categorical & - & - & - & - & $147$\\
            \hline
            Target enrollment & Citeline & Numeric & $3$ & $126$ & $393.03$ & $84965$ & -\\
            \hline
            Drug name & Citeline & Categorical & - & - & - & - & $3418$\\
            \hline
            Inclusion & Citeline & Text & - & - & - & - & - \\
            \hline
            Exclusion & Citeline & Text & - & - & - & - & - \\
            \hline
            Phase\textsuperscript{\textdagger} & DQS & Categorical & - & - & - & - & 8\\
            \hline
            No. site & DQS & Numeric & $1$ & $15$ & $42.56$ & $1140$ & - \\
            \hline
            No. enrolled$^\ast$ & DQS & Numeric & $1$ & $85$ & $292.24$ & $11719$ & - \\
    \hline \hline
    \end{tabular}
    }
\end{table*}


            For the Deep Poisson-Gamma modeling use case described in Section~\ref{sec:method:poissongamma}, we collect the trial-site-level data from IQVIA DQS, which is a more granular version of the data presented previously.  
            There are three additional attributes needed for this use case compared to Section~\ref{sec:dataset}, namely, the~\textit{total number of patients enrolled at trial-site level}, \textit{trial duration}, and \textit{enrollment rate on trial-site level}. 
            The \textit{enrollment duration} is derived from the \textit{trial start date} and \textit{trial completion date} attributes from DQS, while the \textit{enrollment rate on trial-site level} calculated by taking the ratio between the \textit{the total number of patients enrolled at trial-site level} and the difference between the \textit{trial duration} and the \textit{site startup-time}.

        \subsection{Experimental Settings}
        \label{sec:setting}
        \subsubsection{Patient Enrollment Prediction Setting}
            \label{sec:study-level-setting}
            In line with standard procedures, we split the data into three subsets, the training set ($9410$ trials), the development set ($1000$ trials), and the test set($1000$ trials). 
            The dataset is split in such a way that the distributions of number of enrolled patients in these training, development and test sets are similar. 
            We then optimize the model parameters using the training set and select the model exhibiting the best performance on the development set. Subsequently, the evaluation metrics are calculated on the test set to obtain an unbiased assessment of the proposed model. 
            We utilize two widely used metrics for measuring regression performance: the Mean Absolute Error (MAE) and the Coefficient of Determination ($R^2$). 
            $R^2$ is popular metric for regression, however, its reliability can be compromised in case the data contain outliers. 
            Hence, it is beneficial to complement $R^2$ with MAE to quantify the average magnitude of errors. 
            It is important to note that although the model is trained on the logarithmic scale (see Section~\ref{sec:method}), MAE and $R^2$ are calculated on the original scale (i.e., number of participants) of data to maintain interpretability.
            
            We employ both classical machine learning models and advanced language models (PLMs/LLMs) as baselines for performance comparison. For classical models, we use the established gradient boosting models, including XGBoost~\cite{chen2016xgboost} and LightGBM~\cite{ke2017lightgbm}. These models have been known for very high performance on tabular datasets. 
            In order to use these models, encoded attributes are concatenated to create a unified input embedding. 
            For PLMs, we fine-tune BioBERT~\cite{lee2020biobert},  ClinicalBERT~\cite{wang2023optimized}, and Clinical Longformer~\cite{li2022clinical} for regression task; these models take as input the serialized text as discussed in Section~\ref{sec:method:deterministic}. 
            For LLMs, we employ the fine-tuning approach, leveraging Llama2~\cite{touvron2023llama}. 
            Similar to Longformer, Llama2 features an extensive context length of $4096$ tokens, which are sufficiently ample to process long clinical text. 
            Given the limited computational resources, we fine-tune the smallest version of Llama2 (7b) with the following instruction:
            
            \say{
                Analyze the subject enrollment of the following clinical trial enclosed in square brackets. 
                Determine the number of subjects as follows:\\
                \{Clinical Trial Text\}: \{Number of subjects\}.
            }
            Using this format, we prepare the training dataset for LLama2 and fine-tune the model using LoRA technique~\cite{hu2022lora}. The implementation is based on library PEFT\footnote{https://github.com/huggingface/peft}. 
            
            For training the proposed model, we leverage AdamW optimization algorithm~\cite{loshchilov2017decoupled}. We use a small dropout rate of $0.3$ for the category branch but we do not apply dropout regularization on other branches. The batch size is set to $256$. We also use two different learning rates: the same learning rate of $0.0001$ for the input modules, while the learning rate for the rest of the model is $0.001$.

            \subsubsection{Deep Poisson-Gamma Setting}
            \label{sec:setting:poissongamma}
                To maintain the statistical stability and avoid irregularities of enrollment data, we choose to filter the data similar to the setting in Zhong \emph{et al.}~\cite{zhong2024enrollment}. 
                Specifically, we keep only trials with more than $10$ sites and a duration between $6$ and $36$ months.                 
                We then use a subset of the training, development, and test sets described in Section~\ref{sec:dataset} by filtering the trials according to the criteria described in Table~\ref{table:poisson_gamma_data_criteria}. After filtering, the training, development, and test sets contain  $635$, $53$, and $61$ trials, respectively.
                \begin{table*}[t]
    \centering
    \caption{
        Trial selection criteria for the Poisson-Gamma modeling.\\
    }
    \label{table:poisson_gamma_data_criteria}
    \footnotesize
     \begin{tabular}{ c | c } 
         \hline \hline
         \textbf {Attribute} & \textbf {Value} \\
        \hline  \hline
            Trial completion date & Between 2015 and 2025 \\
        \hline 
            Number of sites & More than $10$ sites \\
        \hline
            Trial duration & Between 6 and 36 months\\
        \hline
            Trial phase & II and III\\
        \hline  \hline
    \end{tabular}
\end{table*}

                We follow the same procedure outlined in Section~\ref{sec:study-level-setting}, and introduce two additional metrics, namely, the Median Average Error (MedAE) and the 6-month coverage. The latter represents the percentage of trials for which the actual duration falls within a 6-month window centered around our prediction. For the first approach leveraging our stochastic model described in Section~\ref{sec:method:poissongamma}, we train our stochastic model with a learning rate of $0.0001$ and a batch-size of $128$ for $2048$ epochs. After training our model $\mathcal{F}_{(\mu, \theta)}$, we use it to simulate the enrollment $1024$ times with the Poisson-Gamma model. Each of these simulations produces a predicted duration $\hat{\tau}$. Subsequently, 
                the predicted durations are averaged to obtain our final prediction.

                For the ``filtering-and-fitting'' approach, we use an RMSprop optimizer to find the parameters $\alpha$ and $\lambda$ of our distributions $\Gamma_{\mu,i}$ and $\Gamma_{\theta,i}$ for each trial. The parameters of the optimizer are a learning rate of $0.01$, a batch-size of $128$ and the optimization is carried out across $512$ epochs. The final prediction is obtained in the same way as for the first approach. Moreover, it should be noted that, for the two different approaches, the maximum duration allowed for the prediction is two times the trial duration upper-limit described in Table~\ref{table:poisson_gamma_data_criteria} to avoid unnecessary computation while keeping enough margin for potential errors.
            
        \subsection{Result}
        \label{sec:result}
            \subsubsection{Patient Enrollment Prediction}
            \begin{table*}[t]
    \centering
    \caption{The performance of the proposed method in comparison with baseline models. $R^2$ is the coefficient of determination.
    MAE is the mean absolute error. 
    ($\uparrow$) indicates that higher values reflect better performance while ($\downarrow$) indicates that smaller values correspond to better performance.
    For XGBoost and LightGBM, * indicates that the embeddings of textual attributes are used. 
    Baseline (first row) shows the MAE and $R^2$ calculated between planned and actual numbers.\\
    }
    \label{table:node_classification_result}
    \footnotesize
     \begin{tabular}{ c | c | c } 
         \hline \hline
         Method & $R^2$ ($\uparrow$) & MAE ($\downarrow$) \\
        \hline
        \hline
        Baseline (Target number of patients) & 0.39 & 71.42 \\
        \hline
        XGBoost*~\cite{chen2016xgboost} & 0.70 & 66.74 \\
        \hline
        XGBoost~\cite{chen2016xgboost} & 0.72 & 62.22  \\
        \hline
        LightGBM*~\cite{ke2017lightgbm} & 0.71 & 62.61 \\
        \hline
        LightGBM~\cite{ke2017lightgbm} & 0.76 & 56.84 \\
        \hline
        Llama2~\cite{touvron2023llama} & $0.07$ & $110.98$  \\
        \hline
        ClinicalBERT*~\cite{wang2023optimized} & $0.72$ & $81.08$ \\
        \hline
        BioBERT*~\cite{lee2020biobert} & $\mathbf{0.77}$ & $69.36$ \\
         \hline
        Clinical-Longformer-$1536$*~\cite{li2022clinical} & $0.74$ & $59.88$ \\
        \hline
        Clinical-Longformer-$2048$*~\cite{li2022clinical} & $0.73$ & $59.64$ \\
        \hline\hline
        Our Deterministic Model* & $0.76$ & $\mathbf{51.99}$ \\
        \hline
        Our Probabilistic Model* & $\mathbf{0.77}$ & $55.92$ \\
        \hline 
    \end{tabular}
\end{table*}

            The performance of the proposed method compared to baseline models is presented in~Table~\ref{table:node_classification_result}. Classical ML models perform reasonably good, surpassing the naive baseline of target number of patients, which achieves an $R^2$ of $0.39$ and MAE of $71.42$. It is also worth noting that simply concatenating textual embeddings to attribute embeddings does not generally improve performance. Notably, LightGBM without  textual embeddings achieves top performance among classical ML models with an $R^2$ of $0.76$ and MAE of $56.84$. The PLM fine-tuning approach consistently achieves good $R^2$ scores. Nonetheless, ClinicalBERT and BioBERT do not have good performance on MAE, which is not the case for Clinical Longformer with an MAE of $59.64$. This can be explained by the fact that Longformer has much longer context window compared to ClinicalBERT and BioBERT, which allows it to capture the details of the input clinical text.

            Compared to the baselines, our models demonstrate superior performance in both metrics. Specifically, the deterministic model performs impressively, achieving an $R^2$ score of $0.76$, which places it in the second-highest position among the models evaluated. 
            The model excels in terms of MAE, delivering the best performance with a value of $51.99$ - an improvement of approximately $9$\% over the leading baseline model. 
            The stochastic model, 
            despite its inherent variability, secures the second-best position in MAE and outperforms other models in $R^2$, achieving a score of $0.77$.  
            The superior performance of the proposed deterministic and stochastic models can be attributed to its architecture, which can effectively capture and represent multi-modality information.

            \begin{figure}[t!]
                    \centering
                    \includegraphics[scale=0.37]{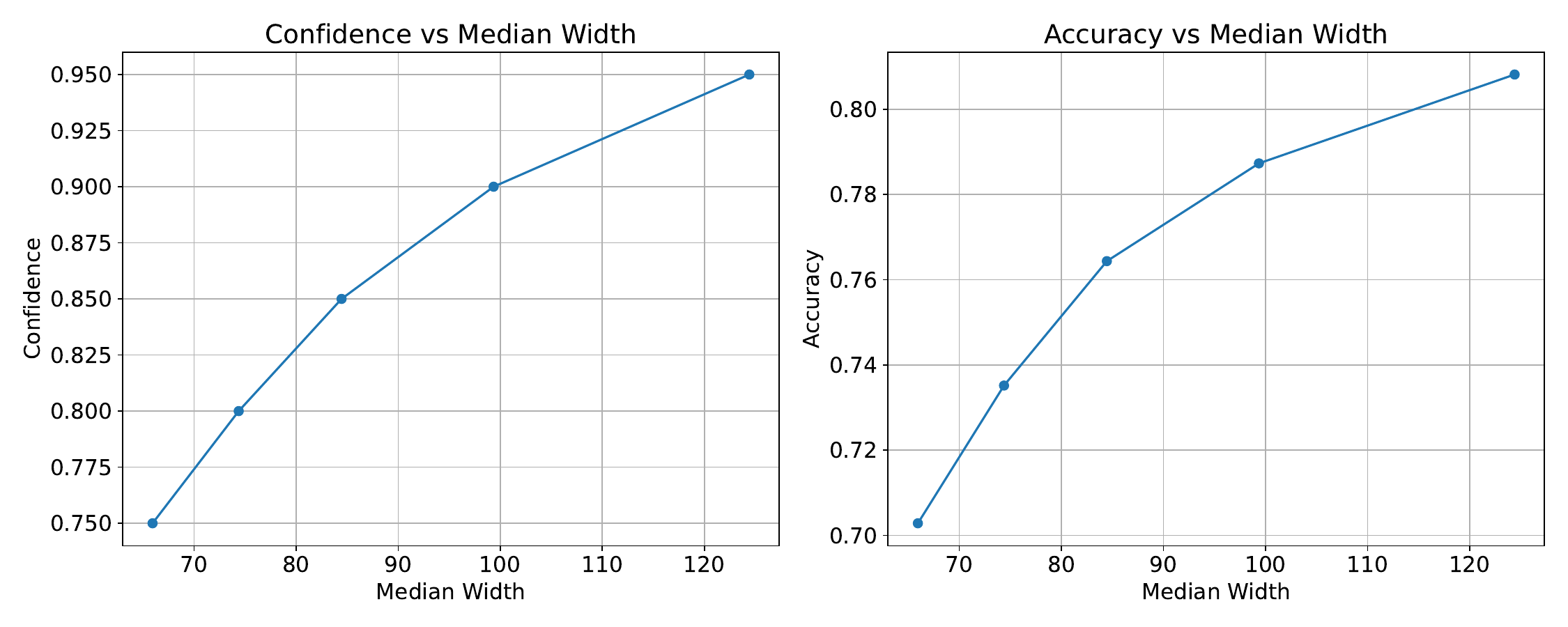}\\
                    \caption{
                    Relationships between interval width, confidence level, and interval accuracy of the stochastic model described in Section~\ref{sec:method:stochastic}.
                    Interval accuracy is the percentage of intervals that contain the actual value of number of patients. 
                    }
                    \label{fig:interval_performance}
            \end{figure}
            
            The proposed stochastic model can be used for range estimation, in addition to point estimation. 
            This ability enables the generation of confidence intervals around enrollment estimates, offering planners a range of plausible outcomes. 
            Specifically, it is possible to predict an interval containing the number of patients with a confidence level of $(1-\alpha)100\%$, where $\alpha$ is the significance level. For instance, choosing $\alpha = 0.1$ leads to a confidence level of $90\%$, a widely used confidence interval. Upon comparison with the actual number of patients, it is observed that $78.73\%$ of ground-truth values fall into the predicted $90\%$ confidence intervals, of which the median value of the interval widths is $99.35$. 
            Adjusting the significance level directly influences the confidence level, interval width, and interval accuracy. This demonstrates the flexibility of the stochastic approach over the deterministic approach. 
            Figure~\ref{fig:interval_performance} illustrates the variation  of confidence level and interval accuracy relative to interval width.

            \subsubsection{Deep Poisson-Gamma}
            \label{sec:result:poissongamma}

\begin{table*}[t]
    \centering
    \caption{
    The performance of the Deep Poisson-Gamma model in comparison with the ``Filtering-and-Fitting'' approach in terms of trial duration prediction. 
    The unit of trial duration is \emph{month}.\\
    }
    \label{table:poisson_gamma_results}
    \footnotesize
     \begin{tabular}{ c | c | c } 
         \hline \hline
          & Filtering-and-Fitting  & Deep Poisson-Gamma  \\
        \hline  \hline
            MAE ($\downarrow$)                    & $9.99$    &  $\mathbf{7.85}$ \\
        \hline
            MedAE ($\downarrow$)                  & $7.59$    &  $\mathbf{6.31}$ \\
        \hline
            6-months coverage [\%] ($\uparrow$)   & $22.95$    &  $\mathbf{26.23}$ \\
        \hline
            Inference time [s] ($\downarrow$)     & $3.83$     &  $\mathbf{0.09}$ \\
        \hline  \hline 
    \end{tabular}
\end{table*}


            As it can be seen in Table~\ref{table:poisson_gamma_results}, our Deep Poisson-Gamma model outperforms the filtering-and-fitting approach on all metrics. 
            More precisely, the former achieves an MAE of $7.85$ months, a two-month error reduction from the latter. 
            Moreover, it can be seen that by leveraging machine learning paradigm, we can significantly reduce inference time. This is due to the fact that for the filtering-and-fitting approach, new distributions need to be fit on similar studies for each trial. This process can be time-consuming as what can be seen as ``training'' is part of the inference process. Our approach leveraging machine learning, allows to separate the training and inference processes and thus greatly reduce the inference time.\\
        
        

    \section{Conclusion}
        \label{sec:conclusion}
        
        Accurately modeling the clinical trial enrollment process, especially in terms of number of patients and enrollment duration, is crucial for clinical trial planning given its potential benefits. 
        In this work, we propose a novel approach to address this challenge. 
        First, we create a novel model to predict the number of patients using trial characteristics. Leveraging the advances in large language models and deep learning, the proposed model is capable of capturing the complexities and nuances of clinical trial documents to effectively make enrollment prediction. 
        We further extend the model by accounting for uncertainties with the Gamma distribution; this way, we naturally connect the statistical approach with the modern machine learning based approach in a unified stochastic model. 
        Using the unified model, it is possible to estimate the trial duration via a Poisson-Gamma model, which covers site-level uncertainties.

        Our experiments show that the proposed method offers superior performance compared to existing approaches in both accuracy and scalability. Our future work will focus on exploring an end-to-end architecture that incorporates large language models directly. In addition, our method is designed to be used at planning phase, when no specific information about sites is available. Hence, we plan to extend this method to consider ongoing trials where the enrollment of patients can be effectively predicted in a near real-time setting to dynamically adjust enrollment forecasts and support adaptive trial management.

\section*{Acknowledgements}
We would like to express our sincere appreciation to James A Southers - Head of Optimization Analytics Ecosystem (OAE, Pfizer) and Melinda M Rottas - Head of Optimization, Analytics and Recruitment Solutions (OARS, Pfizer), for their unwavering support during the preparation of this manuscript. Their thorough reviews and encouragement were instrumental in refining this work and enhancing its overall quality.
\bibliographystyle{abbrv}
\bibliography{enrollment}
\end{document}